\pdfoutput=1
\documentclass{article}

\usepackage{iclr2027_conference,times}
\usepackage[utf8]{inputenc}
\usepackage[T1]{fontenc}
\usepackage{hyperref}
\usepackage{url}
\usepackage{booktabs}
\usepackage{amsfonts}
\usepackage{nicefrac}
\usepackage{microtype}
\usepackage{xcolor}
\usepackage{graphicx}
\usepackage{multirow}
\usepackage{amsmath}
\usepackage{array}

\usepackage{wrapfig}
\usepackage{needspace}

\usepackage{tikz}
\usetikzlibrary{arrows.meta,positioning}
\definecolor{cheapc}{HTML}{0072B2}
\definecolor{expc}{HTML}{D55E00}
\definecolor{routerc}{HTML}{6A3D9A}

\title{Whose Turn Is It? Context-Guided Model\\ Routing for Stateful Agents}

\author{Anikesh Parashar, Avik Dutta, Vashu Chauhan, Sumit Gulwani, Priyanshu Gupta, \\
\textbf{Hosein Hasanbeig, Vu Le, Ashish Tiwari, Jack Williams} \\
Microsoft}

\iclrfinalcopy

\begin{document}

\maketitle
\lhead{Preprint}

\begin{abstract}

Model routing can reduce the cost of LLM applications by assigning simpler requests to cheaper models without sacrificing the quality of the response. However, most routers treat requests as independent. Stateful agents do not operate under this assumption: a user turn may require multiple model and tool calls, its difficulty depends on the state produced by earlier turns, and changing models can sacrifice reuse of the conversational prompt cache. We study zero-shot, cost-aware routing at the user-turn boundary, systematically varying the routing rule, access to session history, candidate models that differ in size, reasoning effort, or model provider, and prompt-cache behavior. We evaluate on multi-turn versions of SheetCopilot and SpreadsheetBench~2, where persistent workbooks and cross-turn dependencies make isolated routing decisions especially brittle. At a matched offloading rate, giving the router session history improves task pass rate over the same router without history, with gains for every model pair; history-aware routing remains within one point of always using the expensive model. We also found that cost depends on execution dynamics rather than token prices alone: the cheaper model’s per-task cost advantage narrows from $6.8\times$ to $3.2\times$ on longer sessions. A closed-form cache analysis further shows that downward switches (expensive to cheap) recover their cache cost within two LLM calls, whereas upward switches do not break even under the evaluated conditions. Together, these results show that agent routing is a sequential decision problem: effective policies must account for what happened earlier in the session and for the asymmetric cost of switching models.

\end{abstract}

\section{Introduction}
\label{sec:intro}

LLM agents interleave model calls with tool execution to complete user tasks. Assigning every call to an expensive frontier model provides strong planning and error recovery, but the cost accrues on every call of every turn of every session. Stronger reasoning is also not uniformly beneficial: extended reasoning can overthink and abandon correct answers while consuming substantially more tokens \citep{zhou-etal-2026-thinking}. Assigning every call to a cheap model reduces cost but fails when a turn requires planning, cross-artifact coordination, or recovery from an earlier error, as smaller models tend to favor quick, locally plausible responses over deliberate reasoning \citep{ranaldi-freitas-2024-aligning}. Model routing exploits this gap by using the cheap model when it suffices and reserving the expensive model for the requests that need it. Throughout this paper, cost reduction is the objective and quality is a constraint: a routing policy is useful if it lowers cost without lowering task success.

Most routers treat requests as independent inputs \citep{chen2023frugalgptuselargelanguage,ong2025routellmlearningroutellms,hu2024routerbenchbenchmarkmultillmrouting}. Deployed agents, however, operate over multi-turn sessions. We use \emph{turn} for one user request together with the agent's response to it, which may involve many LLM calls and tool invocations; a turn is an interaction-level unit, not a single model call. Across turns, routing becomes a sequential problem for four reasons. First, difficulty varies within a session when working on spreadsheets: building a pivot table, changing one of its parameters, and repairing a broken chain of formulas require different capabilities. Second, the current request is often uninterpretable without the preceding trajectory. For instance, ``Apply the same format to Q3'' is simple only if the router knows which format and which sheet, and LLMs are known to lose track as user intent evolves over turns \citep{tack2026llms}. Third, earlier decisions shape later turns: once an expensive model has built the structure of a solution, a cheap model can often extend it by following the traces in its context, so a turn that is hard in isolation can be easy in context. Fourth, switching is not free: providers cache prompt prefixes per model, so a switch forfeits the cached conversation and bills it again as uncached input.

We restrict attention to routing policies that can be deployed as-is, which imposes two constraints. First, routers must be \emph{zero-shot}: their decisions come without routing labels, preference data, or fine-tuning. Candidate models are replaced every few months, and a learned router must be re-supervised for every new pair. In agentic sessions this supervision is especially expensive: labeling one turn requires executing its full tool loop under each candidate from the same state, and because the state at later turns depends on every earlier choice, a $T$-turn session admits up to $2^T$ routed trajectories. Our analysis nonetheless applies to learned routers, which instantiate the same policy class with a trained decision function: the dimensions we study (i.e., what the router observes, when it decides, and in which direction it may switch) do not depend on how that function is obtained. Second, policies must be \emph{cost-aware}. Methods designed to raise quality by spending additional compute, such as sampling several solutions and voting \citep{wang2023selfconsistency}, multi-agent debate \citep{du2024debate}, or verifier-guided best-of-$N$ search \citep{snell2025scaling}, do not target cost reduction, and in agentic sessions, where every sample is a full tool loop, they multiply the cost of each turn. We therefore exclude them.

We study these questions on spreadsheet manipulation domain. Four properties make this domain specifically hard (if not harder than the software-engineering tasks on which agentic systems are most often evaluated \citep{jimenez2024swebench}): (i)~\emph{Two-dimensional, implicitly structured state.} Tables, headers, merged cells, and formatting carry meaning that must be serialized into text, which strains context budgets \citep{dong2024spreadsheetllm}, and LLMs see far less of this structure during pretraining than they see source code. (ii)~\emph{Hidden dependencies.} Formulas form a dataflow graph through cell references rather than through named, importable symbols, so a local edit can silently change values on other sheets. (iii)~\emph{Weak verification.} A repository ships with tests, compilers, and tracebacks that agents use to check their work; a workbook has no executable specification, and a formula over the wrong range returns a plausible number rather than an error. Such silent errors are pervasive even in spreadsheets built by people \citep{panko1998spreadsheet}. (iv)~\emph{End-user intent.} Requests come from end users rather than developers and are short, underspecified, and grounded in the data rather than in a precise issue report. The empirical gap is large: frontier models that lead software-engineering benchmarks (e.g., Claude Opus 4.8 \citep{anthropic2025opus48}) pass at most 31.2\% of our multi-turn SheetCopilot tasks, and on SpreadsheetBench~2 the best of eight frontier models completes 34.9\% of workflows, with debugging accuracy as low as 12\% \citep{zhu2026spreadsheetbench2evaluatingagents}. The same properties make spreadsheets a stringent test of routing: weak verification undermines cascades that decide after inspecting an output, and a persistent workbook makes the difficulty of each turn depend on what earlier turns produced.

We formalize routing at the user-turn boundary. This granularity lies between session-level commitment, which cannot adapt as difficulty changes, and per-call routing, which adds decisions inside the agent loop. We use it to answer four questions: (1) which decision rule to use (an LLM router that assigns whole turns, escalation, or de-escalation); (2) what the router should observe (the current request only, or the full session history); (3) how the answers depend on the candidate pair (differing in model size, reasoning effort, or model provider family); (4) and how switching interacts with prompt caching. We evaluate on multi-turn versions of SheetCopilot \citep{li2023sheetcopilotbringingsoftwareproductivity} and SpreadsheetBench~2 \citep{zhu2026spreadsheetbench2evaluatingagents} and charge router calls and cache effects to every policy.

Our contributions are as follows:
\begin{itemize}
  \item We formulate cost-aware model routing for agents in a multi-turn setting at user-turn granularity, separating what the router observes, how often it decides, and in which direction it may switch; routing without history and routing once per session are special cases.
  \item We build a zero-shot evaluation framework on two multi-turn spreadsheet benchmarks that charges router overhead and prompt-cache effects to each policy, and apply it to candidate pairs that differ in model size, reasoning effort, and model provider.
  \item We show that routing for agents is a sequential decision problem: gains come from conditioning each decision on the session history, and they depend on the direction of switching, on the candidate pair, on cache reuse and depth of a conversation.
\end{itemize}

\section{Problem Formulation}
\label{sec:method}

\begin{figure}[t]
  \centering
\resizebox{\textwidth}{!}{%
\begin{tikzpicture}[
  font=\footnotesize,
  >={Stealth[length=1.7mm,width=1.3mm]},
  box/.style={draw, rounded corners=2pt, line width=0.6pt, align=center, inner sep=2pt},
  flow/.style={->, line width=0.6pt, draw=black!75},
  lbl/.style={font=\scriptsize, text=black!70, inner sep=1.5pt, align=center},
  seg c/.style={fill=cheapc!22},
  seg e/.style={fill=expc!55},
  rowlbl/.style={font=\scriptsize, anchor=east, inner sep=1pt},
  grp/.style={font=\scriptsize\itshape, text=black!55, anchor=west, inner sep=0pt},
]
\node[box, draw=black!60, fill=black!4, minimum width=18mm, minimum height=6.5mm] (u) at (0.9,3.0) {Request $u_t$};
\node[box, draw=black!60, fill=black!8, minimum width=18mm, minimum height=6.5mm, font=\scriptsize] (h) at (0.9,1.5)
  {{\footnotesize History $h_t$}};
\node[box, draw=routerc, fill=routerc!10, line width=0.8pt, minimum width=12mm, minimum height=24mm] (pi) at (4.05,2.25)
  {Router\\ $\pi$\\[2pt]{\scriptsize Escalation:}\\[-2pt]{\scriptsize $m_c$}\\[2pt]{\scriptsize De-escalation:}\\[-2pt]{\scriptsize $m_e$}};
\node[box, draw=cheapc, fill=cheapc!15, minimum width=17mm, minimum height=6.5mm] (mc) at (7.1,3.0) {Cheap $m_c$};
\node[box, draw=expc, fill=expc!15, minimum width=17mm, minimum height=6.5mm] (me) at (7.1,1.5) {Expensive $m_e$};
\node[box, draw=black!75, minimum width=15mm, minimum height=22mm] (agent) at (9.3,2.25)
  {Turn $t$\\[3pt]{\scriptsize LLM calls}\\[-2pt]{\scriptsize $\leftrightarrow$ tools,}\\[-2pt]{\scriptsize edits to}\\[-2pt]{\scriptsize workbook}};
\draw[flow] (u.east) -- node[lbl, above] {no history} (u.east -| pi.west);
\draw[flow, dashed] (h.east) -- node[lbl, above] {with history\\[-2pt]context} (h.east -| pi.west);
\draw[flow, draw=cheapc] (pi.east |- mc.west) -- node[lbl, above, text=cheapc!80!black] {simple} (mc.west);
\draw[flow, draw=expc] (pi.east |- me.west) -- node[lbl, above, text=expc!75!black] {complex /\\[-2pt]fallback} (me.west);
\draw[flow, draw=cheapc] (mc.east) -- (mc.east -| agent.west);
\draw[flow, draw=expc] (me.east) -- (me.east -| agent.west);
\draw[flow, densely dotted, draw=black!55] (u.north) -- ++(0,0.5) -| node[lbl, pos=0.25, above] {no routing: one model for the whole session} (agent.north);
\draw[flow] (agent.south) -- ++(0,-0.8) -| node[lbl, pos=0.25, above] {append turn $t$ $\Rightarrow h_{t+1}$} (h.south);
\def\bx{12.13}\def\bw{0.51}\def\bp{0.60}\def\bh{0.30}
\foreach \k in {1,...,4} \node[font=\scriptsize, text=black!60] at ({\bx+(\k-1)*\bp+\bw/2},3.98) {\k};
\node[rowlbl, text=black!60] at (\bx-0.13,3.98) {turn};
\node[grp] at (10.45,3.62) {no routing};
\node[grp] at (10.45,2.56) {one model per turn};
\node[grp] at (10.45,1.50) {switch within a turn};
\foreach \name/\turns/\y in {
    Expensive-only/{e/e/3,e/e/3,e/e/3,e/e/3}/3.28,
    Cheap-only/{c/c/3,c/c/3,c/c/3,c/c/3}/2.90,
    LLM router/{e/e/3,c/c/3,e/e/3,c/c/3}/2.22,
    Random/{c/c/3,e/e/3,e/e/3,c/c/3}/1.84,
    Escalation/{c/e/2,c/e/3,c/e/1,c/e/3}/1.16,
    De-escalation/{e/c/2,e/c/1,e/c/3,e/c/1}/0.78} {
  \node[rowlbl] at (\bx-0.13,\y) {\name};
  \foreach \fa/\fb/\nf [count=\ti] in \turns {
    \pgfmathsetmacro{\xl}{\bx+(\ti-1)*\bp}
    \begin{scope}
      \clip[rounded corners=1pt] ({\xl},{\y-\bh/2}) rectangle ++(\bw,\bh);
      \foreach \s in {1,2,3} {
        \ifnum\s>\nf\relax\def\m{\fb}\else\def\m{\fa}\fi
        \fill[seg \m] ({\xl+(\s-1)*\bw/3},{\y-\bh/2}) rectangle ++({\bw/3},\bh);
      }
      \foreach \s in {1,2} \draw[white, line width=0.5pt] ({\xl+\s*\bw/3},{\y-\bh/2}) -- ++(0,\bh);
    \end{scope}
    \draw[black!55, line width=0.45pt, rounded corners=1pt] ({\xl},{\y-\bh/2}) rectangle ++(\bw,\bh);
  }
}
\node[font=\small, anchor=north] at (5.0,0.05) {(a) Per-turn routing and state handoff};
\node[font=\small, anchor=north] at (12.33,0.05) {(b) Decision rules over a session};
\end{tikzpicture}}
  \caption{Routing policies for multi-turn agents. \textbf{(a)}~Before turn $t$, a zero-shot LLM router $\pi$ reads the request $u_t$ and, when routing with history context, the serialized history $h_t$ (dashed). It assigns the whole turn to the cheap model $m_c$ (\texttt{simple}) or the expensive model $m_e$ (\texttt{complex}, or on timeout, malformed output, or service failure); under escalation and de-escalation, every turn instead starts on $m_c$ or $m_e$, respectively. Routing without history drops the history input, and without routing, one model serves the whole session (dotted line). The selected model owns every LLM call and tool interaction of the turn, and the runtime appends the resulting messages, tool observations, errors, workbook state, and routing decision to form $h_{t+1}$; models exchange no hidden states or provider-side KV tensors. \textbf{(b)}~Illustrative assignments over a four-turn session.}
  \label{fig:turn-routing}
\end{figure}
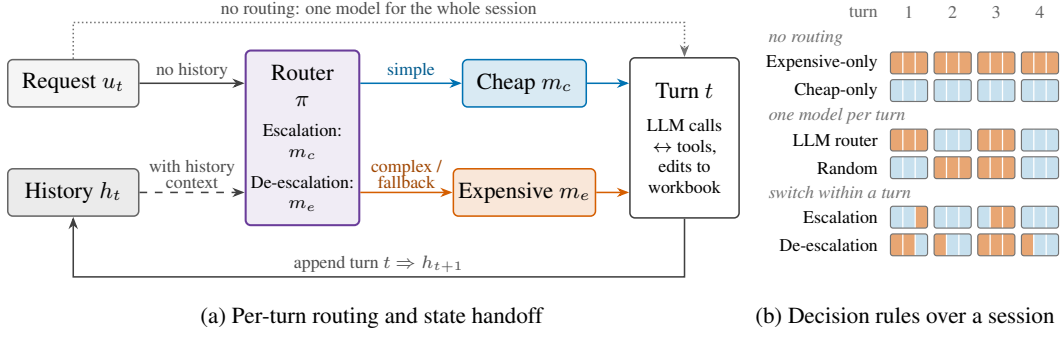

An agentic session consists of user turns $t = 1,\ldots,T$. In each turn, the user sends a request and the agent responds by executing a loop of LLM calls and tool invocations until it returns control to the user. At the start of turn $t$, a router can observe the current request from user $u_t$ and a serialized history $h_t$ containing earlier requests, agent responses, tool calls and observations, execution errors, and prior routing decisions. It selects the model $m_t$:
\begin{equation}
  m_t = \pi(I_t), \qquad I_t \subseteq \{u_t, h_t\}, \qquad m_t \in \mathcal{M} = \{m_c, m_e\},
\end{equation}
where $\pi$ is the router, $I_t$ is the information set at time $t$, $m_c$ is a cheap model and $m_e$ an expensive one. Here \emph{Cheap} means a lower expected cost per turn, through lower per-token prices, fewer reasoning tokens, or both, and usually lower latency. Namely, the two models may differ in size, reasoning effort, or provider. We consider two candidates per session rather than a larger pool. Under per-turn assignment, the selected model handles every LLM call of the turn, while escalation and de-escalation, defined below, may switch models within a turn.

A routing policy has three dimensions. Its \emph{information set} $I_t$ is either the current request alone ($I_t = \{u_t\}$, i.e., routing \emph{without history}) or the request together with the history ($I_t = \{u_t, h_t\}$, routing \emph{with history context}; Figure~\ref{fig:turn-routing}a). Its \emph{decision frequency} specifies the turns at which $\pi$ is invoked. Its \emph{decision rule} specifies how models are assigned. The \emph{LLM router} assigns one model to the whole turn, choosing freely between $m_c$ and $m_e$, and \emph{random} routing does the same while ignoring $I_t$. \emph{Escalation} and \emph{de-escalation} instead switch models within a turn: escalation starts every turn on $m_c$ and hands the rest of the turn to $m_e$ if the router decides to escalate, and de-escalation starts every turn on $m_e$ and hands the rest of the turn to $m_c$ if the router decides to de-escalate (Figure~\ref{fig:turn-routing}b). Routing without history and routing once per session are special cases of this family.

To re-emphasize, we restrict $\pi$ to \emph{zero-shot} policies, whose decision function is given by a prompt and fixed rules rather than learned from routing labels, preference data, or fine-tuning, and to \emph{cost-aware} policies, whose expected cost can fall below that of assigning $m_e$ to every turn. A learned router would instantiate the same policy class with a trained decision function, and thus, our analysis concerns the structure of the policy, which is independent of how that function is obtained.

For task $i$, let $Y_i\in\{0,1\}$ denote end-to-end success and let $C_i$ denote total inference cost, including router overhead and prompt-cache effects. We evaluate a routing policy $\pi$ by the pair
\begin{equation}
  \left(\mathbb{E}[Y_i],\, \mathbb{E}[C_i]\right),
\end{equation}
and compare policies by their position on the quality--cost plane. A policy Pareto-dominates another if it achieves at least as high a pass rate at no greater cost, with a strict improvement in at least one dimension. Because our objective is cost reduction, the central question is whether a policy lowers $\mathbb{E}[C_i]$ relative to the expensive-only policy without lowering $\mathbb{E}[Y_i]$.

\textbf{Per-turn Routing.} The LLM router is a lightweight LLM prompted as a zero-shot binary classifier, invoked exactly once after each user request and before the turn executes (Figure~\ref{fig:turn-routing}a). Given $(u_t,h_t)$, it emits a structured labels, mapped to $m_c$ or $m_e$. Under the LLM router, the assignment remains fixed for the entire turn: the router is not reinvoked between the selected model's internal LLM calls or tool interactions. The turn ends only when the agent returns a response and gives control to the user. The agent runtime owns the persistent session and workbook state. Models therefore continue through serialized runtime state; they do not exchange hidden states or provider-side KV tensors.

\textbf{Escalation and De-escalation.} These rules move the decision inside the turn and give it to the working model instead of a separate router. Under \emph{escalation}, every turn starts on $m_c$, which handoffs to $m_e$ when it judges the current request too hard for itself. The call passes the rest of the turn to $m_e$, which continues from the same transcript, tools, and workbook state. \emph{De-escalation} mirrors this rule.

\textbf{Random Routing.} The random router is a baseline for switching without a decision signal. Before every turn, it assigns the whole turn to $m_c$ or $m_e$ at random, independently of $I_t$ and of earlier assignments, and the selected model then runs the turn exactly as under the LLM router.

\section{Experimental Setup}
\label{sec:exp_setup}

We evaluate the routing policies on spreadsheet manipulation. In this setup the workbook is persistent, mutable state that later turns build on, and its hidden dependencies and weak verification signals make both the agent's task and the router's decision quite complex.

\textbf{Benchmarks.} We convert single-turn benchmarks into multi-turn conversations by decomposing each query into individual subtasks. Each turn operates on the workbook state resulting from the preceding subtasks, with the sequence collectively reproducing the outcome of the original single-turn request at the end of the conversation:

\textit{SheetCopilot.}
SheetCopilot contains 221 spreadsheet tasks with ordered subtasks \citep{li2023sheetcopilotbringingsoftwareproductivity}. We retain the 112 tasks with multiple subtasks and issue each subtask as a separate user turn in the original order. The resulting conversations average 2.88 turns, with a maximum of seven turns.

\textit{SpreadsheetBench~2.}
SpreadsheetBench~2 covers more complex spreadsheet tasks in finance and business domain \citep{zhu2026spreadsheetbench2evaluatingagents}. We exclude debugging tasks due to limited contextual carry-over, and convert 221 end-to-end workflows into multi-turn conversations following \citet{tack2026llms}. The resulting conversations average 4.76 turns, with a maximum of six turns.

\textbf{Metrics.}
We evaluate over the scope of the entire conversation rather than turn by turn.
Per-turn metrics alone do not reveal the compounding effect of a multi-turn exchange, so conversation-level scoring is what exposes the net impact of model switching on final task correctness and whether it saves cost.
For task $i$, $Y_i=1$ if the final workbook at the end of the conversation passes the benchmark's own grading procedure against the gold answer, and we report the pass rate, the mean of $Y_i$ over tasks, as our estimate of $\mathbb{E}[Y_i]$.
The cost $C_i$ sums the cost of every LLM call in the conversation at the prices in Table~\ref{tab:models}, billing cached input tokens at the cached rate and all other input tokens at the uncached rate, so cache effects enter through the token counts; we report its mean per task-run as our estimate of $\mathbb{E}[C_i]$.

\textbf{Agent and candidate models.}
We evaluate all routing policies within a generic spreadsheet-agent harness that uses the same agent, tool definitions, system prompt, and execution limits in every condition. The agent reads and edits the workbook through tool calls and runs each user turn as a loop of LLM calls and tool invocations until it returns control to the user; the runtime, not the model, owns the session and workbook state. Candidate pairs are chosen to vary one factor at a time: model size within a model provider; reasoning effort within a single model; and model provider. All models run at medium reasoning effort unless stated otherwise, and GPT-5.6 Luna serves as the router LLM in every condition. Table~\ref{tab:models} in Appendix~\ref{app:cost} lists the models and prices.

We study three candidate pairs, each varying one factor: GPT-5.6 Sol versus GPT-5.6 Luna (model size within a model provider), GPT-5.6 Sol versus Claude Haiku 4.5 (a cheap model from another model provider), and Claude Opus 4.8 at high versus low reasoning effort (reasoning effort within a single model). All policies share one agent runtime: the system prompt, tools, execution limits, serialized trajectory, candidate pair, and task order are fixed, and a policy may change only the model assignment. Table~\ref{tab:policies} defines each policy at the level needed to reproduce its decisions and state transitions.

\begin{table}[t]
  \centering
  \caption{Routing policies compared in our evaluation}
  \label{tab:policies}
  \small
  \begin{tabular}{>{\centering\arraybackslash}p{2.1cm}>{\centering\arraybackslash}p{1.5cm}>{\centering\arraybackslash}p{2.3cm}>{\arraybackslash}p{6.4cm}}
    \toprule
    \textbf{Policy} & \textbf{When} & \textbf{Decision signal} & \textbf{Assignment and state behavior} \\
    \midrule
    Expensive-only / Cheap-only & Fixed & None & Assign $m_e$ / $m_c$ to every turn; the quality--cost endpoints. \\
    LLM router, no history & Per-turn & $u_t$ & Assign the entire multi-call loop for turn $t$ to either model from the current request alone; adopts the information set of request-level routers \citep{ding2024hybridllmcostefficientqualityaware,ong2025routellmlearningroutellms}. \\
    LLM router, with history & Per-turn & $(u_t,h_t)$ & Same assignment, conditioned on the serialized session history. DialRouter and MTRouter are the closest history-aware precedents but are learned and route at different granularities \citep{zhang2026myopicselectionlonghorizonawareness,zhang2026mtroutercostawaremultiturnllm}. \\
    Escalation & Within turn & $(u_t,h_t)$ & Start every turn on $m_c$; hand the rest of the turn to $m_e$ if the router decides to escalate. \\
    De-escalation & Within turn & $(u_t,h_t)$ & Start every turn on $m_e$; hand the rest of the turn to $m_c$ if the router decides to de-escalate. \\
    Random & Per-turn & None & Select a model at random before every turn; controls for switching without a decision signal \citep{zhang2026mtroutercostawaremultiturnllm}. \\
    \bottomrule
  \end{tabular}
\end{table}

\section{Evaluation}
\label{sec:evaluation}

We organize the evaluation around four research questions:
\begin{description}
  \item[RQ1 (Decision rule).] How do the LLM router, escalation, de-escalation, and random routing trade off cost and pass rate, and how do they compare with the fixed-model baselines?
  \item[RQ2 (Router context).] Does conditioning on the session history improve routing over routing without history?
  \item[RQ3 (Candidate pair).] How do these trade-offs depend on whether the candidates differ in model size, reasoning effort, or model provider?
  \item[RQ4 (Caching).] How does switching affect prompt caching and its efficiency?
\end{description}

\subsection{RQ1: Decision rule and comparison with baselines}
\label{sec:rq1}

\begin{figure}[t]
\centering
\includegraphics[width=\textwidth]{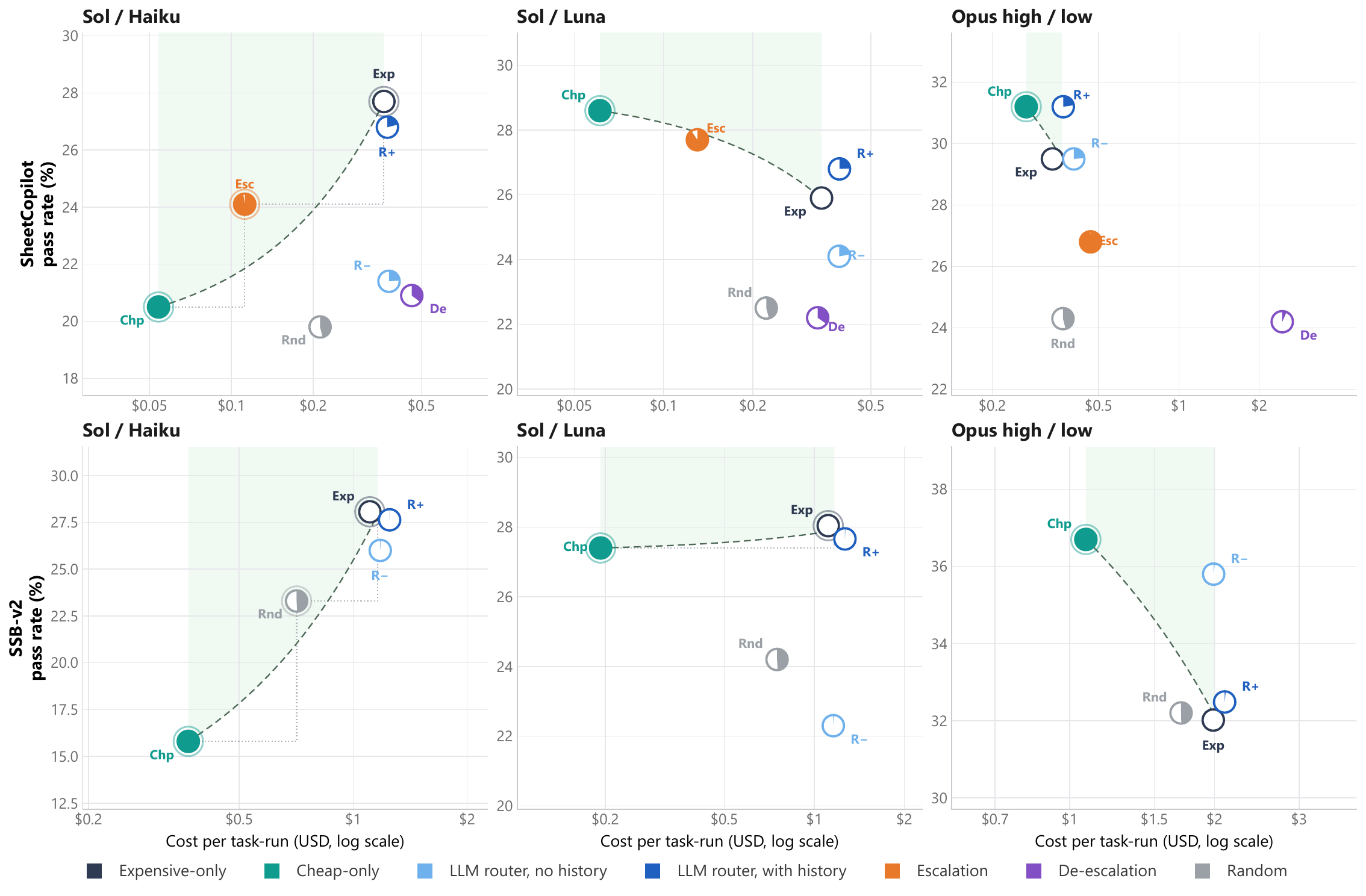}
\caption{Cost--quality tradeoff of the decision rules in Tables~\ref{tab:main}
and~\ref{tab:ssb2}; rows are benchmarks, columns are candidate pairs, axes are
scaled per panel. Colour gives the rule (legend), abbreviated at each marker as
Exp, Chp, R$-$, R$+$, Esc, De, and Rnd. The filled wedge is $m_c$: empty $=0\%$,
half $=50\%$, solid $=100\%$. The dashed curve traces all random mixes of the
two pure candidates, and the shaded band above it is where a rule beats such a
mix. A bold ring marks a Pareto-optimal rule and the dotted staircase joins
them. Coincident markers are nudged apart for legibility.}
\label{fig:baseline-comparison}
\end{figure}

Figure~\ref{fig:baseline-comparison} reports, for every decision rule and candidate pair on SheetCopilot and SpreadsheetBench 2, our estimates of $\mathbb{E}[Y_i]$ (pass rate) and $\mathbb{E}[C_i]$ (cost per task-run), together with the share of turns served by the cheap model.

\textbf{The LLM router with history largely preserves expensive-model quality.} On SheetCopilot, expensive-only and cheap-only achieve average pass rates of 27.7\% and 26.8\%, respectively, while the LLM router with history achieves 28.3\% while routing 21.1--24.2\% of turns to the cheap model. Its pass rate is within 0.9 points of expensive-only for all three pairs and exceeds it for Sol/Luna and Opus high/low. On SSB-2, the router uses the cheap model on only 0.4--1.7\% of turns and achieves an average pass rate of 29.3\%, compared with 29.3\% for expensive-only and 26.6\% for cheap-only. Thus, on both benchmarks, the history-based router maintains pass rates close to the expensive-only baseline while making relatively few downward switches. Whether this translates to cost savings requires per-model token accounting; for Opus high/low, the only pair with exact routed costs, the router costs \$0.369 versus \$0.365 for expensive-only on SheetCopilot and \$2.080 versus \$2.009 on SSB-2.

\textbf{Escalation mostly retains the cheap model.} On SheetCopilot, 90.4--99.4\% of turns remain on the cheap model, with escalation rates of 2.2\%, 9.6\%, and 0.6\% for Sol/Haiku, Sol/Luna, and Opus high/low, respectively. Its pass rates are 24.1\%, 27.7\%, and 26.8\%, compared with 20.5\%, 28.6\%, and 31.2\% for cheap-only. On SSB-2, 99.0\%, 83.4\%, and 92.4\% of turns remain on the cheap model, corresponding to escalation rates of 1.0\%, 16.6\%, and 7.6\%. These gains over cheap-only come at substantially higher cost, particularly for Opus high/low (\$2.456 vs.\ \$1.081).

\textbf{De-escalation and random routing provide weaker trade-offs.} On SheetCopilot, de-escalation routes 34.7--35.5\% of turns to the cheap model for the Sol pairs and only 6.7\% for Opus high/low, with pass rates of 20.9--24.2\%. It trails the fixed-model baselines on average, and its Opus high/low cost (\$2.440) is substantially higher than either fixed baseline.
Random routing sends approximately 46\% of turns to the cheap model on SheetCopilot and 49\% on SSB-2, but achieves lower pass rates than both fixed-model endpoints across all three pairs on both benchmarks. Its mean pass rate is 22.2\% on SheetCopilot and 26.6\% on SSB-2. These results show that routing without a task-dependent signal does not consistently improve the cost--quality trade-off over simply using one of the fixed models.

\subsection{RQ2: Router context and continuity across switches}
\label{sec:switching}

\textbf{Routing with and without history.} We isolate the router's information set by comparing the LLM router with history context, $I_t=\{u_t,h_t\}$, against the same router without history, which also decides before every turn but observes only $u_t$, excluding prior dialogue, tool observations, execution errors, and routing decisions (Table~\ref{tab:policies}). Both policies make the same number of decisions, so the comparison isolates the contribution of the trajectory $h_t$.

The two routers send nearly the same share of turns to the cheap model, 23.3\% without the history and 22.7\% with it on average, yet the history-aware router passes 28.3\% of tasks against 25.0\% without history. The 3.3-point gain holds in every pair: 5.4 points for Sol/Haiku, 2.7 for Sol/Luna, and 1.7 for Opus high/low. Because the offloading rate is matched, the gain comes from \emph{which} turns are offloaded rather than how many. Without $h_t$, follow-up requests such as parameter changes or references to previously created workbook objects appear deceptively simple or ambiguous, and the router misallocates them; it needs the trajectory to determine what a request refers to, what has already been attempted, and whether the current turn requires recovery. The gain is largest for Sol/Haiku, the pair with the widest quality gap between its candidates (7.2 points), where a misrouted turn costs the most. A turn is therefore not an independent routing instance: its appropriate assignment depends on the trajectory leading to it, which makes routing a sequential decision problem. On SpreadsheetBench~2, both routers overwhelmingly retain the expensive model, with the cheap-model share at only 1.5--1.7\% without history and 0.4--1.7\% with history; the history-aware router achieves 29.3\% average pass rate versus 28.0\% without history, indicating a smaller and less consistent benefit when offloading is rare. A limitation of this comparison is that SheetCopilot follow-ups tend to be direct continuations of the first request $u_1$, which already indicates the domain of the session.

\subsection{RQ3: Dependence on the candidate pair}
\label{sec:rq3}

\textbf{The router ignores the pair.} The history-aware router sends 21--24\% of turns to the cheap model in every pair (Table~\ref{tab:main}), which is further minimal for ssb2, although the pairs differ sharply: Haiku trails Sol by 7.2 points, whereas Luna and Opus at low effort match or exceed their expensive counterparts on SheetCopilot. The router thus responds to the perceived difficulty of a turn rather than to the capability gap between the candidates, and its pass rate follows the pair: it ends 0.9 points below expensive-only for Sol/Haiku but 0.9 and 1.7 points above it for Sol/Luna and Opus high/low, where offloading costs little quality. A router informed of the pair could offload far more turns where the cheap candidate is as accurate as the expensive one. In those pairs the cheap model alone is also the better choice: Luna passes 28.6\% of tasks at \$0.061 per task-run against 26.8\% for the router with history, and Opus at low effort matches the router's 31.2\% at \$0.268 rather than \$0.369 (Table~\ref{tab:main}). Routing within a pair pays off only if the expensive candidate is worth its price, so choosing the pair matters as much as routing within it.

\begin{figure}[t]
\centering
\includegraphics[width=\textwidth]{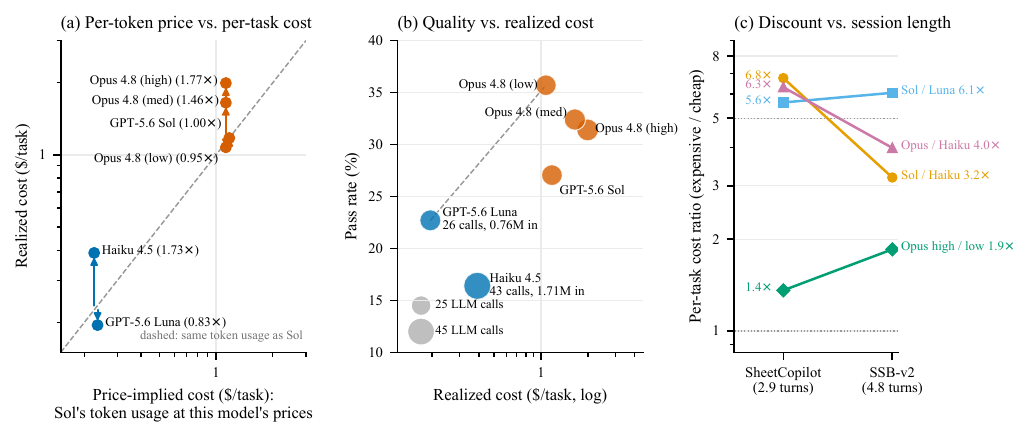}
\caption{Per-token price versus per-task cost. \textbf{(a, b)}~SpreadsheetBench~2 single-model runs (repeated Sol runs averaged). \textbf{(a)}~Arrows show the cost implied by GPT-5.6 Sol's token usage at each model's prices versus realized cost $\mathbb{E}[C_i]$, with the ratio in parentheses. \textbf{(b)}~Pass rate $\mathbb{E}[Y_i]$ versus realized cost; marker area denotes LLM calls per conversation and the dashed line the Pareto frontier. Haiku 4.5 makes about 1.6$\times$ more calls and uses 1.7$\times$ more input tokens, reducing its per-task discount to 3.2$\times$ versus Sol; Luna costs about half as much as Haiku while passing 6 points more. Opus 4.8 at low effort is both cheaper and more accurate than at high effort. \textbf{(c)}~Ratio of expensive-to-cheap $\mathbb{E}[C_i]$ on SheetCopilot versus SpreadsheetBench~2; dotted lines show list-price ratios. Haiku's discount shrinks with session length, while Luna's remains stable.}
\label{fig:cost-per-task}
\end{figure}

\textbf{Cheap per token is not cheap per task.} On the longer SpreadsheetBench~2 sessions, list price is a poor proxy for the cost of a task, $\mathbb{E}[C_i]$ (Figure~\ref{fig:cost-per-task}). Haiku 4.5 is priced at one fifth of GPT-5.6 Sol per input token, but it takes about 41 LLM calls and 1.6M input tokens per conversation, against about 25 calls and 0.9M tokens for the other models. It therefore costs only 3.2$\times$ less than Sol per task (\$0.37 versus \$1.18) while passing 16.4\% of tasks rather than 27.1\%. GPT-5.6 Luna, at a similar list price, keeps Sol's call budget (26 calls) and costs about half as much as Haiku (\$0.19) with a 6.3-point higher pass rate. The cheap candidate should therefore be chosen by its realized cost per task on the target workload, not by its list price.

\textbf{Multi-turn sessions change the discount.} Comparing the two benchmarks shows how this gap scales with the session (Figure~\ref{fig:cost-per-task}c). On SheetCopilot's shorter sessions (2.9 turns on average), Haiku uses about as many calls as Sol (13.6 versus 14.1) and is 6.8$\times$ cheaper per task, above the 5$\times$ list-price ratio. On the longer SpreadsheetBench~2 sessions (4.8 turns), its discount falls to 3.2$\times$. Luna's discount holds at 5.6--6.1$\times$, and the cost gap between Opus 4.8 at high and low effort widens from 1.4$\times$ to 1.9$\times$. Every LLM call re-reads the accumulated session, and the mean context per call roughly doubles, from about 18K tokens on SheetCopilot to about 38K on SpreadsheetBench~2, so extra calls cost more the longer the session. A model's per-task cost therefore depends on how its call budget grows with the session, which short, single-turn evaluations do not reveal. The two benchmarks also differ in workbook size and domain, so this comparison is descriptive.

\subsection{RQ4: Interaction with prompt caching}
\label{sec:rq4}

\textbf{Where switching costs arise.} Cache reuse differs between switched and unswitched turns. When consecutive turns use the same model, the next turn can reuse both the stable system-and-tool prefix and the model-specific conversation trajectory accumulated so far. After a switch, the newly selected model can still reuse its system-and-tool prefix, which remains warm across sessions because this prefix is fixed for that model in our deployment. However, it cannot reuse the trajectory cached by the previous model; the conversation beyond the shared base prefix is processed as uncached input until the new model builds its own cached prefix. The first LLM call within each turn therefore receives \emph{at least} a stable cached-prefix length regardless of which model handled the previous turn.

\begin{wrapfigure}{r}{0.38\columnwidth}
    \centering
    \includegraphics[width=\linewidth]{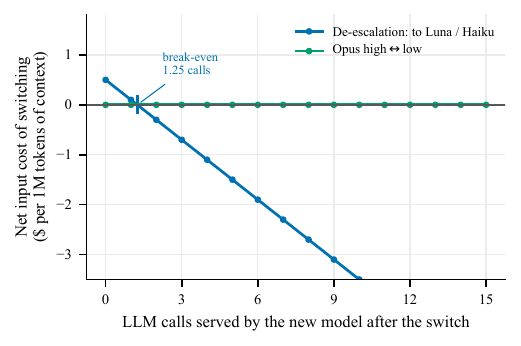}
    \caption{
        Net cost of switching from Sol to Luna / Haiku per million uncached input tokens, as a function of LLM calls. The switch becomes cost-effective when net cost crosses zero.
    }
    \label{fig:switch-breakeven}
\end{wrapfigure}

Switching therefore incurs two distinct costs, both part of $C_i$: the router call and the higher input cost caused by losing trajectory-cache reuse. The second appears in the token counts of the calls after a switch; the first is charged separately. Conversely, the fixed-model baseline receives the most cache-favorable treatment because it is assumed to preserve the growing trajectory cache across every turn in a task. This assumption makes our cost comparison conservative for routed policies: in a deployment where session disconnects, cache expiration, or cold restarts invalidate the trajectory prefix, fixed-model runs would also pay for uncached trajectory tokens and their cost advantage from uninterrupted cache reuse would shrink.

\textbf{When does a switch pay off?} Let $L$ be the length of the conversation beyond the warm system-and-tool prefix at the moment the router switches from $m_e$ to $m_c$, and let $p^{m}_{\text{in}}$ and $p^{m}_{\text{cache}}$ be model $m$'s uncached and cached input prices. To first order, the first call on $m_c$ reads $L$ as uncached input instead of from $m_e$'s cache, which adds $L\,(p^{c}_{\text{in}} - p^{e}_{\text{cache}})$ to $C_i$, and every later call on $m_c$ reads $L$ from $m_c$'s cache, which saves $L\,(p^{e}_{\text{cache}} - p^{c}_{\text{cache}})$. The switch therefore pays for itself after
\begin{equation}
  k^{*} = \frac{p^{c}_{\text{in}} - p^{e}_{\text{cache}}}{p^{e}_{\text{cache}} - p^{c}_{\text{cache}}}
\end{equation}
calls on $m_c$, independently of $L$ (Figure~\ref{fig:switch-breakeven}). At our prices, $k^{*} = 1.25$ for Sol$\to$Luna and Sol$\to$Haiku, so a downward switch recovers its cost within one turn if it sticks. Because $k^{*}$ grows quickly as the cheap model's uncached price approaches the expensive model's cached price, a candidate that is only moderately cheaper may never repay a switch within a session. An upward switch ($m_c\to m_e$) instead costs $L\,(p^{e}_{\text{in}} - p^{c}_{\text{cache}})$, about \$4.9 per million tokens of context, ten times the \$0.5 of a downward switch to Luna or Haiku, and it never pays back on input because every later call reads the more expensive cache. Since $L$ grows with every turn, switching up late is the expensive direction: at 75K tokens of context, one upward switch costs about as much as an entire SheetCopilot task on Sol, whereas switching down early is nearly free. This yields a testable design principle: switching should be \emph{sticky} and \emph{aware of context length}, switching down early and staying there, and switching up only when the expected quality gain outweighs a cost that grows with the session.

\section{Related Work}

\textbf{Cost-aware model routing.} Most routing work selects a model for each independent request to balance quality, cost, or latency. Learned routers predict from preference data or performance labels whether a cheap model suffices \citep{ding2024hybridllmcostefficientqualityaware,ong2025routellmlearningroutellms,hu2024routerbenchbenchmarkmultillmrouting,mei2025omnirouterbudgetperformancecontrollable} or how much test-time compute to allocate \citep{ding2025bestrouteadaptivellmrouting}, with later work addressing changing model pools and online adaptation \citep{jitkrittum2025universalmodelroutingefficient,wang2025mixllmdynamicroutingmixed}. Zero-shot alternatives derive the decision from confidence estimates without training \citep{nguyen2026zeroshotconfidenceestimationsmall}. Cascades run the cheap model first and escalate after inspecting its output \citep{chen2023frugalgptuselargelanguage,aggarwal2025automixautomaticallymixinglanguage,dekoninck2025unifiedapproachroutingcascading,miao2026selective}. Quality-oriented test-time methods such as self-consistency, multi-agent debate, and verifier-guided best-of-$N$ search \citep{wang2023selfconsistency,du2024debate,snell2025scaling} spend additional compute to raise accuracy and are complementary to our cost-reduction objective. In all of these methods, the decision unit is an independent request or generation rather than a turn in an evolving agent session.

\textbf{Multi-turn and agentic routing.} History-aware routing has only recently been studied. DialRouter conditions each choice on the preceding user--assistant dialogue in question answering, with one LLM call per turn, and learns long-horizon behavior from simulated MCTS trajectories \citep{zhang2026myopicselectionlonghorizonawareness}. MTRouter uses task, action, observation, and error history in tool-using agents but learns to select a model for each individual LLM invocation \citep{zhang2026mtroutercostawaremultiturnllm}. HyDRA instead retains its selected model across ordinary turns and reroutes after context compaction or summarization \citep{garg2026hydrahybriddynamicrouting}. We study zero-shot routing at the intermediate user-turn granularity: one model owns the complete multi-call tool loop for a request, and the next model continues from the serialized dialogue, tool observations, errors, and artifact state. We additionally examine the direction of switching and its interaction with prompt caching.

\textbf{Spreadsheet agents.} SheetCopilot introduced LLM agents that manipulate spreadsheets through atomic actions \citep{li2023sheetcopilotbringingsoftwareproductivity}. SpreadsheetBench evaluates real-world manipulation requests collected from online forums \citep{ma2024spreadsheetbench}, SpreadsheetBench~2 evaluates end-to-end business workflows over large multi-sheet workbooks \citep{zhu2026spreadsheetbench2evaluatingagents}, and SpreadsheetLLM studies how to encode two-dimensional workbooks for LLMs \citep{dong2024spreadsheetllm}. Agentic evaluation has otherwise concentrated on software engineering \citep{jimenez2024swebench,openai2024swebenchverified}, where executable tests provide the verification signal that spreadsheets lack. To our knowledge, model routing has not previously been studied for spreadsheet agents.

\section{Limitations and Conclusion}

\textbf{Limitations.}
We study zero-shot, cost-aware policies by design. Learned routers and quality-oriented methods such as voting could improve routing or pass rates, but are complementary to our analysis. Our sessions average nearly three turns, with trajectories up to seven turns; longer trajectories, independent tasks, and cross-sheet interactions remain future work. Finally, decomposing annotated subtasks into sequential turns approximates natural interaction but does not capture the full diversity of organic dialogues.

\textbf{Conclusion.} We studied zero-shot, cost-aware model routing for multi-turn LLM agents on spreadsheet manipulation, a domain that is harder than software engineering in the ways that matter for routing. Because spreadsheet sessions alternate between construction, inspection, parameterization, and repair, committing to one model for an entire session is unnecessarily coarse, yet re-deciding every turn from the current request alone is worse. Our results characterize routing as a sequential decision problem: at the same offloading rate, a router that reads the session history passes 3.3 points more tasks than one that sees only the current request and keeps the quality of the expensive model, whereas self-assessed escalation and de-escalation and random switching do not. Which pair is routed matters as much as how: when the cheap candidate is as accurate as the expensive one, it dominates every routed policy of its pair. Prompt caching adds a cost to every switch that grows with the session, so switching should be sticky and aware of context length.

\clearpage
\bibliography{citation}
\bibliographystyle{iclr2027_conference}
\newpage

\appendix

\section{Model pricing and cost accounting}
\label{app:cost}

Table~\ref{tab:models} reports the prices used to calculate inference costs. Prices were recorded on August 29, 2026, from the providers' published pricing documentation. We preserve these prices for all reported experiments even if provider prices subsequently change. All amounts are in USD per one million tokens and exclude taxes (if applicable) and account-specific discounts.

\begin{table}[ht]
  \centering
  \caption{Candidate models and prices per one million tokens.}
  \label{tab:models}
  \begin{tabular}{@{}llrrr@{}}
    \toprule
    Role & Model & Input & Cached & Output \\
    \midrule
    Expensive & GPT-5.6 Sol & \$5.00 & \$0.50 & \$30.00 \\
    Cheap, router & GPT-5.6 Luna & \$1.00 & \$0.10 & \$6.00 \\
    Cheap & Haiku 4.5 & \$1.00 & \$0.10 & \$5.00 \\
    Expensive (high), cheap (low) & Opus 4.8 & \$5.00 & \$0.50 & \$25.00 \\
    \bottomrule
  \end{tabular}
\end{table}

\section{Cost-Quality Trade-off}
\label{appx:cost-quality-tradeoff}

\begin{table}[h]
\centering
\caption{Decision rules on SheetCopilot for the three candidate pairs. Pass: conversation-level pass rate, our estimate of $\mathbb{E}[Y_i]$ (\%). $m_c$: share of turns served by the cheap model (\%); for escalation, turns never escalated; for de-escalation, turns handed to $m_c$. Cost: mean cost per task-run in dollars, our estimate of $\mathbb{E}[C_i]$, excluding the router call. LLM-router costs are exact only for Opus high/low, whose candidates share prices. Escalation, de-escalation, and random routing ran in a separate harness. Tasks: 112, except random (111) and de-escalation (110 for Sol/Haiku, 99 otherwise).}
\label{tab:main}
\small
\setlength{\tabcolsep}{4pt}
\begin{tabular}{@{}l*{3}{ccc}@{}}
\toprule
 & \multicolumn{3}{c}{\textbf{Sol / Haiku}} & \multicolumn{3}{c}{\textbf{Sol / Luna}} & \multicolumn{3}{c}{\textbf{Opus high / low}} \\
\cmidrule(lr){2-4}\cmidrule(lr){5-7}\cmidrule(l){8-10}
\textbf{Policy} & Pass & $m_c$ & Cost & Pass & $m_c$ & Cost & Pass & $m_c$ & Cost \\
\midrule
Expensive-only          & 27.7 & 0    & 0.364 & 25.9 & 0    & 0.341 & 29.5 & 0    & 0.365 \\
Cheap-only              & 20.5 & 100  & 0.054 & 28.6 & 100  & 0.061 & 31.2 & 100  & 0.268 \\
\midrule
LLM router, no history    & 21.4 & 23.9 & 0.380 & 24.1 & 22.4 & 0.391 & 29.5 & 23.6 & 0.372 \\
LLM router, with history  & 26.8 & 21.1 & 0.375 & 26.8 & 24.2 & 0.392 & 31.2 & 22.7 & 0.369 \\
Escalation              & 24.1 & 97.8 & 0.112 & 27.7 & 90.4 & 0.130 & 26.8 & 99.4 & 0.467 \\
De-escalation           & 20.9 & 35.5 & 0.461 & 22.2 & 34.7 & 0.331 & 24.2 & 6.7  & 2.440 \\
Random                  & 19.8 & 46.0 & 0.212 & 22.5 & 46.0 & 0.222 & 24.3 & 46.0 & 0.368 \\
\bottomrule
\end{tabular}
\end{table}

\begin{table}[h]
\centering
\caption{Decision rules on SSB-2 for the three candidate pairs. Pass: conversation-level pass rate, our estimate of $\mathbb{E}[Y_i]$ (\%). $m_c$: share of turns served by the cheap model (\%); for escalation, turns never escalated; for de-escalation, turns handed to $m_c$. Cost: mean cost per task-run in dollars, our estimate of $\mathbb{E}[C_i]$, excluding the router call. LLM-router costs are exact only for Opus high/low, whose candidates share prices. Escalation, de-escalation, and random routing ran in a separate harness.}
\label{tab:ssb2}
\small
\setlength{\tabcolsep}{4pt}
\begin{tabular}{@{}l*{3}{ccc}@{}}
\toprule
 & \multicolumn{3}{c}{\textbf{Sol / Haiku}} & \multicolumn{3}{c}{\textbf{Sol / Luna}} & \multicolumn{3}{c}{\textbf{Opus high / low}} \\
\cmidrule(lr){2-4}\cmidrule(lr){5-7}\cmidrule(l){8-10}
\textbf{Policy} & Pass & $m_c$ & Cost & Pass & $m_c$ & Cost & Pass & $m_c$ & Cost \\
\midrule
Expensive-only            & 27.9 & 0    & 1.159 & 27.9 & 0    & 1.168 & 32.1 & 0    & 2.009 \\
Cheap-only                & 15.8 & 100  & 0.367 & 27.4 & 100  & 0.193 & 36.7 & 100  & 1.081 \\
\midrule
LLM router, no history    & 26.0 & 1.5  & 1.179 & 22.3 & 1.5  & 1.158 & 35.8 & 1.7  & 1.993 \\
LLM router, with history  & 27.8 & 0.4  & 1.192 & 27.8 & 0.7  & 1.208 & 32.4 & 1.7  & 2.080 \\
Random                    & 23.3 & 49.2 & 0.709 & 24.2 & 49.1 & 0.751 & 32.2 & 49.6 & 1.705 \\
\bottomrule
\end{tabular}
\end{table}

\end{document}